\documentclass{article}

\usepackage{microtype}
\usepackage{graphicx}
\usepackage{subfigure}
\usepackage{booktabs} 

\usepackage{hyperref}

\usepackage{amsmath}
\usepackage{amssymb}
\usepackage{mathtools}
\usepackage{amsthm}

\usepackage{hyperref}
\usepackage{url}
\usepackage{wrapfig}
\usepackage{enumitem}
\usepackage{multirow} 
\usepackage{tcolorbox}
\usepackage{colortbl}    
\usepackage{booktabs}       
\usepackage{amsfonts}       
\usepackage{nicefrac}       
\usepackage{microtype}      
\usepackage{pifont}
\usepackage{listings}
\usepackage{arydshln}
\usepackage[misc]{ifsym}
\usepackage{tabularx}
\usepackage{array}

\usepackage[accepted]{icml2025}
\usepackage{amsmath}
\usepackage{amssymb}
\usepackage{mathtools}
\usepackage{amsthm}
\usepackage{wasysym}
\usepackage{fontawesome}

\usepackage[capitalize,noabbrev]{cleveref}
\theoremstyle{plain}

\theoremstyle{definition}

\theoremstyle{remark}

\usepackage[textsize=tiny]{todonotes}

\newcommand{\name}{{\color{black} PhysiAgent}}

\begin{document}

\twocolumn[
\icmltitle{PhysiAgent: An Embodied Agent Framework in Physical World}



\icmlsetsymbol{equal}{*}
\icmlsetsymbol{lead}{$\dagger$}
\icmlsetsymbol{intern}{$\ddagger$}

\icmlsetsymbol{corr}{\Letter}
\begin{icmlauthorlist}
\icmlauthor{Zhihao Wang}{equal,lead,intern,air,pku}
\icmlauthor{Jianxiong Li}{equal,air}
\icmlauthor{Jinliang Zheng}{equal,air}
\icmlauthor{Wencong Zhang}{intern,air}
\icmlauthor{Dongxiu Liu}{intern,air}
\icmlauthor{Yinan Zheng}{air}
\icmlauthor{Haoyi Niu}{intern,air,ucb}
\icmlauthor{Junzhi Yu}{corr,pku}
\icmlauthor{Xianyuan Zhan}{corr,air}
\end{icmlauthorlist}

\icmlaffiliation{air}{AIR, Tsinghua University}
\icmlaffiliation{pku}{Peking University}
\icmlaffiliation{ucb}{University of California, Berkeley}

\icmlcorrespondingauthor{Zhihao Wang}{zh1haowang@stu.pku.edu.cn}
\icmlcorrespondingauthor{Junzhi Yu}{yujunzhi@pku.edu.cn}
\icmlcorrespondingauthor{Xianyuan Zhan}{zhanxianyuan@mail.tsinghua.edu.cn}

\icmlkeywords{Machine Learning, ICML}

\vskip 0.3in
]



\printAffiliationsAndNotice{\icmlEqualContribution \icmlProjectLead \icmlIntern} 

\begin{abstract}
Vision-Language-Action (VLA) models have achieved notable success but often struggle with limited generalizations.
To address this, integrating generalized Vision-Language Models (VLMs) as assistants to VLAs has emerged as a popular solution. However, current approaches often combine these models in rigid, sequential structures: using VLMs primarily for high-level scene understanding and task planning, and VLAs merely as executors of lower-level actions, leading to 
ineffective collaboration and poor grounding challenges. In this paper, we propose an embodied agent framework, \textbf{\name}, tailored to operate effectively in physical environments. By incorporating monitor, memory, self-reflection mechanisms, and lightweight off-the-shelf toolboxes, \name~offers an autonomous scaffolding framework to prompt VLMs to organize different components based on real-time proficiency feedback from VLAs to maximally exploit VLAs' capabilities.
Experimental results demonstrate significant improvements in task-solving performance on complex real-world robotic tasks, showcasing effective self-regulation of VLMs, coherent tool collaboration, and adaptive evolution of the framework during execution. \name\ makes practical and pioneering efforts to integrate VLMs and VLAs, effectively grounding embodied agent frameworks in real-world settings.
\end{abstract}

\section{Introduction}
\label{sec:introduction}

    Recent advancements in Vison-Language-Action (VLA) models 
    have demonstrated impressive progress in diverse applications, such as real-world tabletop manipulation~\cite{kim2024openvla,zheng2025universal,chi2024diffusionpolicy,black2024pi_0}, language-guided navigation~\cite{zhang2024uni,zhang2024navid} and human-robot interactions~\cite{shi2024yell,shi2025hi,intelligence2025pi05visionlanguageactionmodelopenworld}. These models are generally built upon large-scale, high-quality labeled datasets and trained using an end-to-end paradigm~\cite{black2024pi_0,intelligence2025pi05visionlanguageactionmodelopenworld}. Despite their strong potential across various applications, VLA models often struggle to generalize to unseen scenarios, tasks, and embodiments~\cite{chen2024roviaug}, which hinders their practical deployment. Simply scaling data and models is highly ineffective to overcome the generalization challenge, considering the high cost of data collection and the diminishing returns in performance as data volume increases~\cite{lin2024data}.

    To address this issue, researchers turn to leveraging powerful Visual Language Models (VLMs). The extensive common knowledge within these models allows for enhanced visual reasoning, task decomposition, and proprioceptive understanding~\cite{hurst2024gpt}. However, existing methods often employ VLMs solely as high-level planners with a rigid, disjoint language interface, resulting in inadequate grounding~\cite{ahn2022can,huang2022inner}. 
        While a recent trend involves unifying VLMs and VLAs into single embodied foundation models via joint training, the substantial data requirements and the risk of catastrophic forgetting of general knowledge after fine-tuning constrain the scalability and potential of these methods~\cite{zheng2025universal,black2024pi_0,reussflower}. 
        
        In contrast to the embodied AI community's struggle with rigid model integration, the Large Language Model (LLM) Agent community has demonstrated remarkable success in modular collaboration. By leveraging carefully designed scaffolding, LLM agents integrate multiple specialized components to exhibit behaviors that go beyond the capabilities of individual models. These components, each with a distinct function, interact autonomously under the guidance of high-level rules, resulting in organically cooperative behaviors and impressive performance in virtual tasks such as gaming~\cite{wang2023voyager,wang2023describe,wang2024jarvis,qin2024mp5}, human simulation~\cite{yang2024oasis,park2023generative}, and disease diagnosis~\cite{li2024agent}. Despite the success of LLM Agents in virtual environments, the potential of LLM agent-style modular coordination for real-world physical systems remains largely unexplored.

   In this paper, we introduce \textbf{\name}, a novel training-free embodied agent framework designed to enable the seamless integration and deployment of VLMs and VLAs in the physical world, addressing the longstanding challenge of generalization. Central to \name\ is a unified, fully automated scaffolding that conceptualizes the entire embodied system as a cohesive intelligent agent, capable of self-regulation and dynamic adaptation across its internal components.
    Unlike conventional hierarchical embodied systems with rigid and feedforward architectures, \name\ incorporates agent memory, monitor and a self-reflective mechanism to establish an adaptive feedback loop that supports deeper grounding and coordination between arbitrary VLMs and VLAs. Specifically, VLMs continuously learn the proficiency of VLAs across various instructions and skills through ongoing autonomous interaction, enabling adaptive strategy refinement without human intervention.
    To further enhance the system's capacity for grounded reasoning and precise control, \name\ integrates thinking, perception, and control assistance modules alongside VLMs and VLAs. These components, implemented via lightweight rule-based programs or well-trained neural models, can be invoked on demand at minimal computational cost. All modules, including VLMs, VLAs, and assistance mechanisms, are orchestrated by the automated scaffolding within an asynchronous execution paradigm, allowing them to operate concurrently and interact fluidly. This framework, which naturally aligns with the ongoing evolution of VLMs, enables them to engage in autonomous reasoning, self-reflection, and interaction—unlocking the full potential of VLAs and auxiliary modules, and ultimately achieving improved generalization and performance in real-world environments.

    To evaluate the effectiveness of our framework, we conducted real-world experiments in a tabletop manipulation scenario, using a single robotic arm as the primary platform for physical interaction. Off-the-shelf VLMs and VLAs were seamlessly integrated into our proposed scaffolding and prompted to complete a variety of manipulation tasks.
    Notably, \name\ exhibited agent-like behaviors during deployment, such as self-reflection and adaptive evolution throughout task execution. These emergent capabilities arose not from any single module, but from the dynamic interactions and feedback among internal components. Quantitatively, our framework achieved a substantial improvement in task success rates while maintaining an acceptable operational frequency, underscoring the practical advantages and robustness of our approach.

    In summary, our contributions are threefold: 1) we propose \name, a training-free, modular embodied agent framework that flexibly integrates VLMs and VLAs for real-world deployment; 2) we make a novel step in bringing the agent paradigm—traditionally explored in language or simulated domains—into the physical world, equipping VLMs with real-world perception and tool-using capabilities;  3) we validate our framework in real-world robotic manipulation tasks, demonstrating emergent self-reflection and significant improvements in task performance.

    \begin{figure*}[t]
    \label{fig:comparison}
    \centering
    \includegraphics[width=1.00\textwidth]{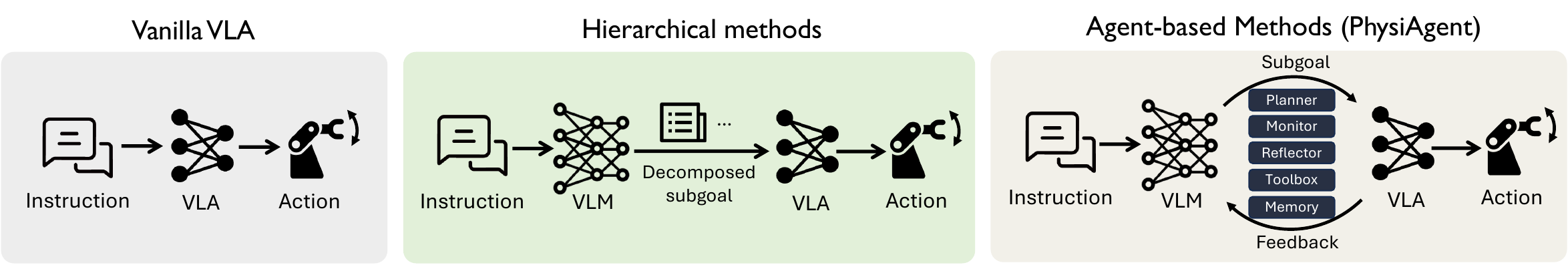}
    \caption{Comparison between (a) vanilla VLA methods, (b) hierarchical embodied AI methods, and (c) agent-based methods. Unlike vanilla VLA and hierarchical approaches, agent-based methods enable coherent and dynamic interaction between VLMs and VLAs.}
    \vspace{-5pt}
    \end{figure*}

\section{Preliminary}
\label{sec:preliminary}
Embodied AI aims to develop models capable of following arbitrary human instructions and interacting with the physical world to accomplish tasks. We denote $\mathcal{L}$ as the language instruction space, $\mathcal{O}$ as the observation space, and $\mathcal{U}$ as the control signal space.
Given an instruction $l \in \mathcal{L}$ and a sequence of observation history up to time step $t$, $o_{\leq t} \subset \mathcal{O}$,
the embodied AI model $f$ aims to produce control signal $u_t \subset \mathcal{U}$ to solve the task described by $l$. The trained model $f$ can be formulated as follows:
\[
u_t = f(o_{\leq t},\ l),
\]
\textbf{Vanilla Vision-Language-Action Models} typically implement $f$ using a unified neural policy, denoted as $\pi_\theta(u_t|o_{\leq t}, l)$, where $\theta$ represents the trainable parameters. Learning this policy purely in an end-to-end paradigm from demonstrations imposes a significant demand on labeled data and often struggles to generalize across diverse tasks and embodiments.

\textbf{Hierarchical methods} extend this formulation by introducing an intermediate variable $z$, which serves as an abstract representation to bridge high-level instructions and low-level control. Under this framework, the policy is factorized as:
\[
\pi_{\theta,\phi}(u_t|o_{\leq t}, l) = \pi_{\theta}(u_t|o_{\leq t}, z_t)\cdot\pi_{\phi}(z_t|o_{\leq t}, l),
\]
where $\pi_\phi$ denotes high-level VLMs and $z_t$ may correspond to a textual or visual subgoal or a latent representation encoding task-relevant intent. While this structure introduces effective modularity, $z_t$ is often rigidly defined or learned without explicit constraints, leading to weak coordination between the two sub-policies and limited robustness in complex environments.

\textbf{Agent systems} are designed to solve complex grounding tasks by decomposing them into sub-tasks and orchestrating their execution through a carefully constructed scaffolding. Unlike hierarchical embodied methods that rely on a naively defined intermediate variable $z_t$, agent systems typically enable information exchange within a closed-loop architecture, supporting self-reflection and adaptive behavior across components. However, building agent systems that can operate effectively in the physical world remains a non-trivial challenge, due to the high demands of 3D egocentric reasoning and robust physical grounding.


\section{Autonomous Embodied Agent Framework}
\label{section:approach}
To address the limitations of the rigid interactions between VLMs and VLAs in existing approaches, we introduce \name. The central to \name\ is an autonomous scaffolding mechanism that integrates VLMs and VLAs into a unified agent system, allowing VLMs to dynamically select and utilize different tools based on VLAs' proficiencies, thus fully unleashing the capabilities of VLA execution. We provide an overview of \name\ in Section~\ref{subsec:overview}, detail its core components as an agent system and key considerations for deployment in real-world scenarios in Section~\ref{subsec:watcher}.

\subsection{PhysiAgent Overview}
\label{subsec:overview}

\begin{figure*}[t]
\centering
\includegraphics[width=1.00\textwidth]{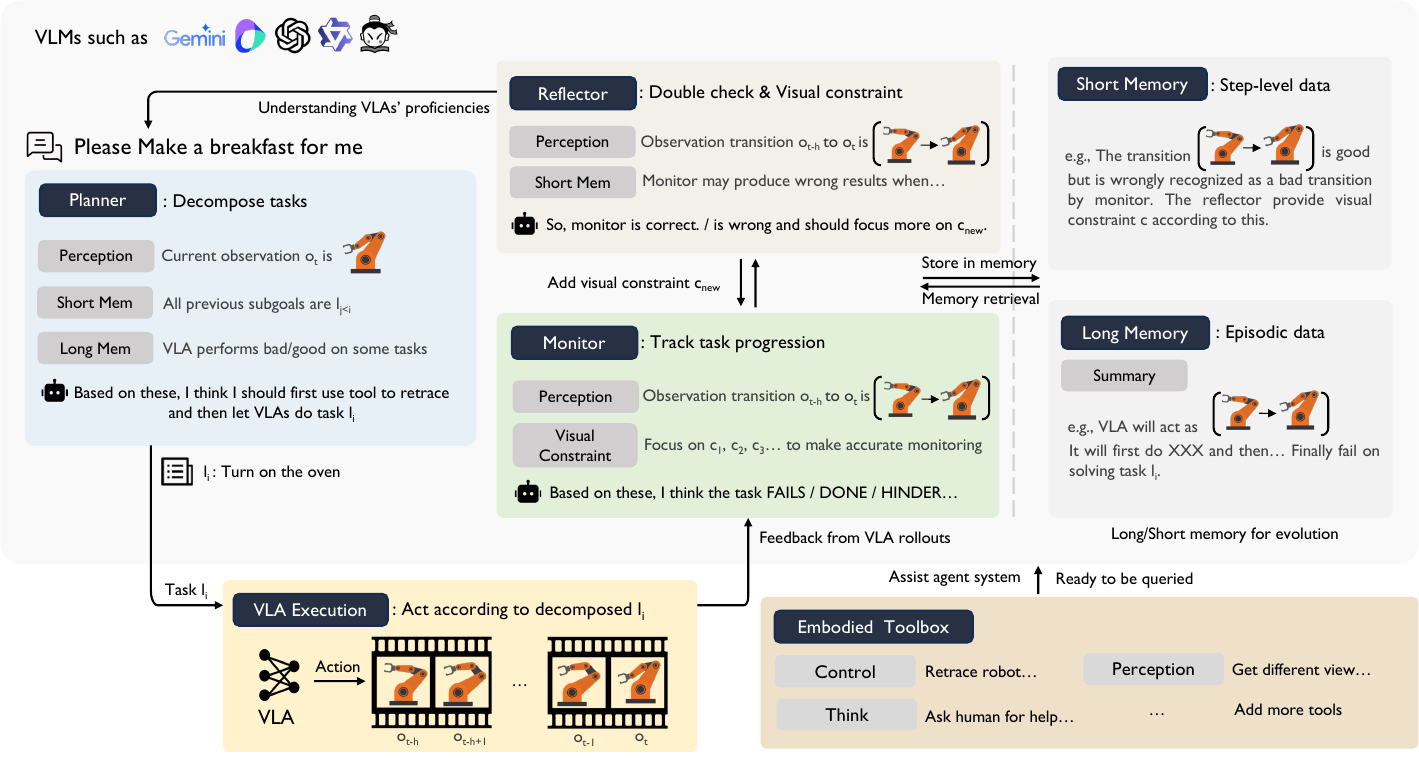}
\caption{The workflow of \name~for executing a given task. The \textit{Planner} converts high-level task requests into executable language instructions for the low-level VLA to perform. Simultaneously, the \textit{Monitor} tracks the VLA’s progress as it follows these instructions. In parallel, the \textit{Reflector} evaluates the monitor’s outputs, generating visual constraints to correct any misjudgments. These reflection results provide insight into the VLA’s proficiency and serve as valuable context for the \textit{Planner} in the next round of task decomposition.
To support continual adaptation, \textit{Short Memory} and \textit{Long Memory} are maintained by capturing step-level and episodic execution data, respectively, informing future planning and reflection through memory retrieval. Additionally, the \textit{Embodied Toolbox} provides a suite of tools for perception, control, and reasoning, enabling the agent system to observe the physical world, command the robot, or seek human assistance, acting as an effective runtime assistant that enhances system robustness and adaptability.
}
\label{fig:outline}
\end{figure*}

As discussed in Section~\ref{sec:preliminary}, hierarchical methods have limitations due to their simplistic integration of VLMs and VLAs. These methods rely on intermediate subgoals or latent variables ($z_t$), resulting in a linear, feedforward flow from instruction to action execution. This approach often produces intermediate representations ($z_t$) inadequately grounded or poorly optimized for VLAs, as VLMs fail to sufficiently consider VLAs' specific execution capabilities.  
To overcome these limitations, \name\ integrates components like monitoring, self-reflection, and memory
inspired by their remarkable success of agents in enhancing multi-component collaboration~\cite{wang2023voyager,wang2023describe,wang2024jarvis,qin2024mp5}. These components encourage VLMs to dynamically understand and adapt to VLAs' proficiencies by iteratively refining intermediate representations based on execution feedback, as illustrated in Figure~\ref{fig:outline}. 

Specifically, \name\ consists of five key components: \textit{Planner}, \textit{Monitor}, \textit{Reflector}, \textit{Memory}, and \textit{Toolbox}. The primary function of the \textit{Planner} is to decompose raw language instructions ($l$) into actionable intermediate instructions ($l_i$) suitable for VLAs. 
\begin{equation}
    \text{Planner}: l_i=F_p(o_t,l_{j<i},l)\\ 
    \label{equ:planner}
\end{equation}
To enable dynamic responsiveness to VLAs' real-time capabilities, we introduce the \textit{Monitor}, which continuously tracks VLA execution progress $\mathcal{P}$. Recognizing that the \textit{Monitor} may occasionally produce inaccurate evaluations due to inherent VLM limitations, we incorporate a \textit{Reflector}. The \textit{Reflector} serves as a verification layer to improve monitoring accuracy. Additionally, the \textit{Memory} component records the overall agent interaction histories in $\mathcal{M}$, providing the \textit{Reflector} and \textit{Planner} with relevant context to enhance its reasoning capabilities and understand VLAs' proficiencies. Lastly, to ensure applicability in real-world physical systems, we introduce an easily implementable yet powerful \textit{Toolbox}, which offers various practical tools to improve the overall system performance.


Consequently, the information flow within \name\ is bidirectional: forward from VLMs to VLAs, and backward from VLAs' behavior to VLMs, allowing VLMs to adjust their outputs in response to real-time feedback. Next, we will elaborate on the details of our introduced \textit{Monitor}, \textit{Reflector}, \textit{Memory} and \textit{Toolbox}, as well as considerations for deploying the systems in physical world.

\subsection{Grounding Enhancement via Monitor, Reflection and Memory}
\label{subsec:watcher}

Existing agent frameworks can operate well in digital world, but we found directly applying them in real-world robotic platforms remains highly suboptimal. The key challenge behind this huge gap is the visual-language grounding challenges. Most existing agents operate purely in text-based environments~\cite{wang2024survey,xi2025rise} with no grounding issues. While some frameworks incorporate visual information~\cite{qin2024mp5,nguyen2024gui,zhang2024large}, they are often limited to simplified settings like GUI elements~\cite{nguyen2024gui}, clean simulator visuals~\cite{qin2024mp5}, or fully textual-describable visual inputs~\cite{fan2022minedojo, achiam2023gpt}. Therefore, how to process and grounding real-world visual inputs is the key challenge we try to resolve.

\textbf{Grounding Equipment via Monitor}. The \textit{Monitor} $F_m$ addresses the grounding challenges by enhancing the VLM's ability to assess task progress. Unlike most LLM-based agents that rely on a single observation frame $o_t$, \textit{Monitor} uses multiple frames to evaluate VLA execution. However, feeding all past frames $o_{\leq t}$ is computationally expensive.
To balance efficiency and effectiveness, we use two successive frames $(o_t, o_{t-h})$, selected by a sliding window of size $h$. Prior works~\cite{li2024decisionnce, ye2025latent, bruce2024genie} have shown that meaningful semantic progress can be extracted from differences between such frame pairs, offering a lightweight yet informative approach.
\begin{align}
   & \text{Monitor}:\notag \\ &p_t=F_m(o_t,o_{t-h},l_i,\mathcal{C}), \notag \\
   &p_t\in\mathcal{P}:=\{{\rm Hinder}, {\rm Ongoing}, {\rm Failure}, {\rm Done}\}
\end{align}
The \textit{Monitor} $F_m$ evaluates whether the transition from $o_{t-h}$ to $o_t$ reflects positive progress toward the instruction $l_i$. While $p_t$ could be a continuous score indicating progress, generating precise fine-grained values remains challenging~\cite{ma2025vision}. To improve reliability and interpretability, we instead prompt VLMs to produce discrete progression flags: $\{{\rm Hinder}, {\rm Prompting}, {\rm Failure}, {\rm Done}\}$. These simplified stages make the monitoring results both more accurate and easier for humans to understand.

Careful readers may notice that although our simplification can simplify the problem and reduce errors, \textit{Monitor} may still produce incorrect progression flags due to the inherent limitations of current VLMs. To address this, we introduce \textit{Reflection} that serves as both a double-checker and an enhanced reasoning module. It outputs reflection results $c$, which are then fed back into the \textit{Monitor} to improve its grounding accuracy. We will detail this mechanism next.

\begin{figure*}[t]
\centering
\includegraphics[width=1.00\textwidth]{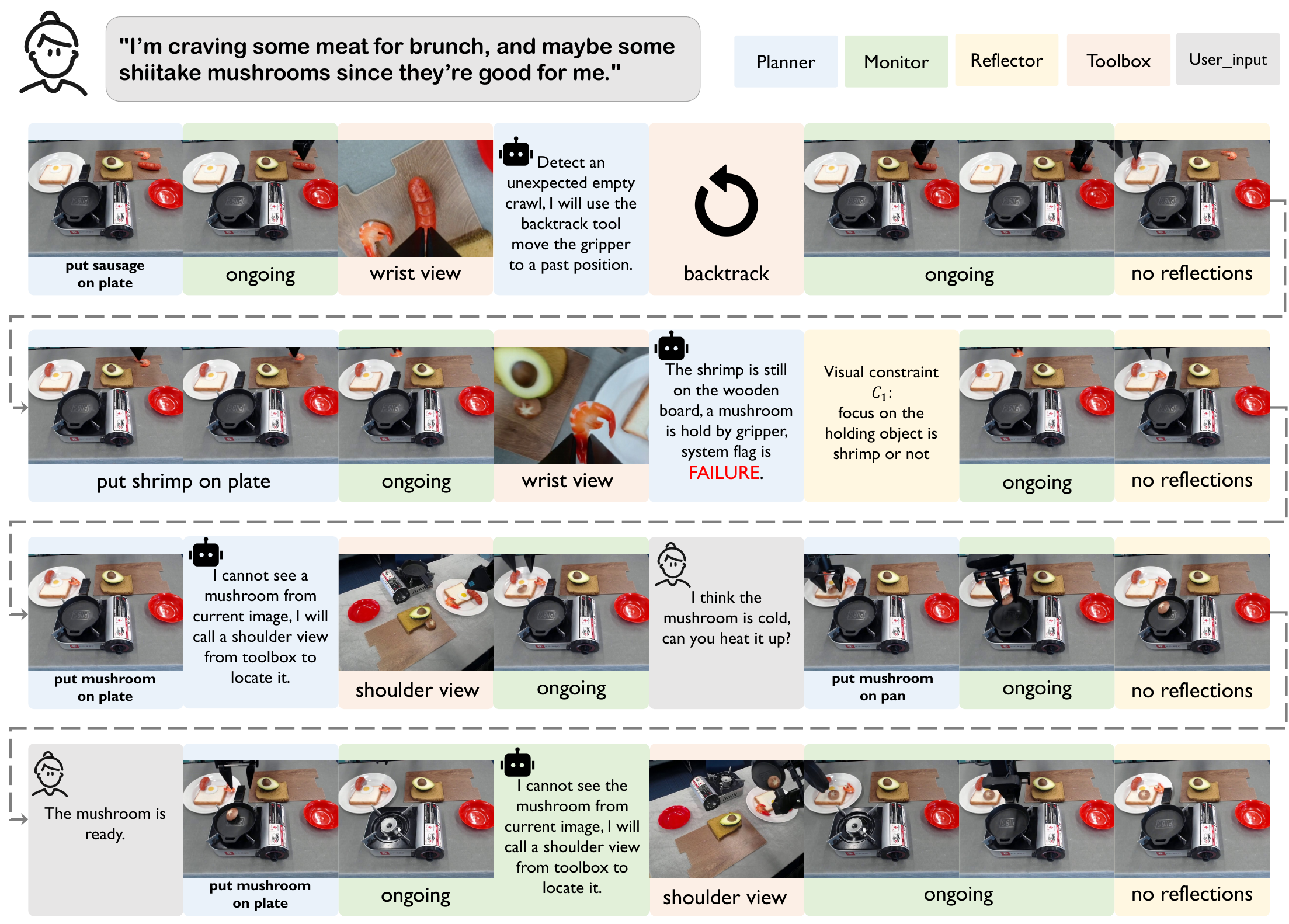}
\caption{An illustrative example of \name~workflow. To accomplish a complex task, \name~coordinates its components in a unified and adaptive manner. Additionally, it can incorporate human prompts to refine its planning when needed, highlighting its potential as a practical and robust framework for real-world embodied agent systems.}
\label{fig:example}
\end{figure*}

\begin{figure*}[t]
\centering
\includegraphics[width=0.9\textwidth]{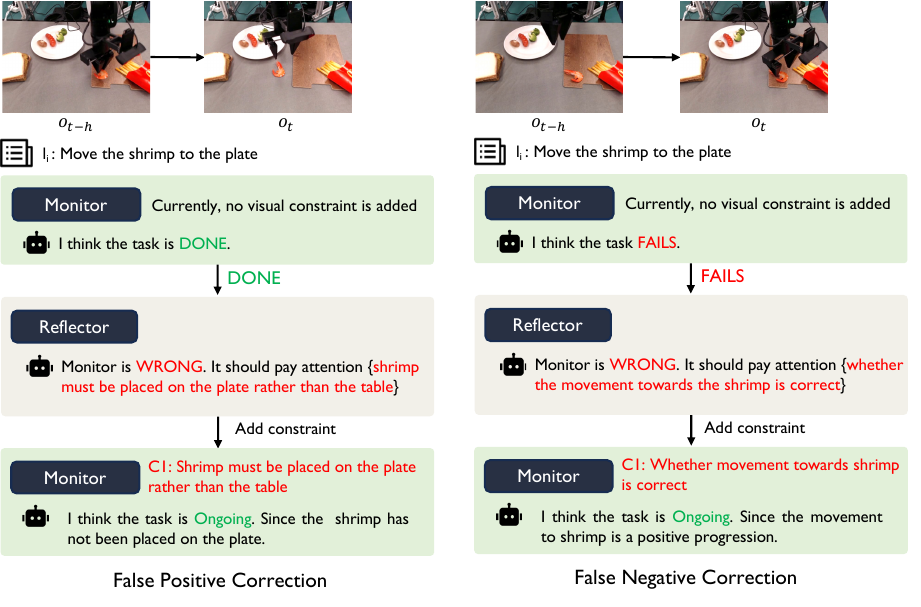}
\caption{Examples of \name's visual constraint workflow. \textit{1) False Positive Correction}. The shrimp is placed on the table, but the task is incorrectly marked as \textit{DONE}. In response, the \textit{Reflector} generates a visual constraint instructing the \textit{Monitor} to more carefully assess the shrimp’s position to prevent similar mistakes. \textit{2) False Negative Correction}. The robot arm correctly moves toward the shrimp, but the action is wrongly flagged as \textit{FAILS}. The \textit{Reflector} addresses this by generating a visual constraint indicating that such motion should be carefully evaluated, improving future evaluations.} 
\label{fig:watcher}
\end{figure*}

\textbf{Grounding Enhancement via Reflection}. To verify the \textit{Monitor}'s evaluation, the \textit{Reflector} $F_r$ primarily takes as input the progression flags $p_t$, visual observations $o_{t-h}$ and $o_t$, and the instruction $l_i$. It cross-checks for inconsistencies between the visual transition ($o_{t-h}, o_t$) and the predicted flag $p_t$. If an inconsistency is detected, the \textit{Reflector} identifies the failure mode and generates a corresponding \textit{visual constraint} $c$, which is stored in a constraint buffer $\mathcal{C}$ to guide the \textit{Monitor} in avoiding similar errors in the future. The formulation is as follows:
\begin{equation}
    \text{Reflector}: c = F_{r}(o_t, o_{t-h}, l_i, p_t), c\rightarrow\mathcal{C}
    \label{equ:reflector}
\end{equation}
For example, in the task “Move the shrimp to the plate” (Figure~\ref{fig:watcher}), the shrimp is placed near the plate but remains on the tabletop. The \textit{Monitor} incorrectly outputs a \textit{DONE} flag, failing to capture the subtle positional error—resulting in a false positive error. The \textit{Reflector} identifies this failure mode and generates a constraint $c_1$: “Shrimp must be placed on the plate rather than the table,” encouraging more precise spatial awareness in future evaluations.

Through repeated interaction between \textit{Monitoring} and \textit{Reflection}, the system accumulates constraints in $\mathcal{C}$, allowing the \textit{Monitor} to adapt and improve grounding over time. This enables our framework to enhance task understanding in a fully training-free manner, offering a practical and deployable solution for embodied agents in real-world settings.

\textbf{Evolution via Memory}. While \textit{Monitor}'s capability improves through accumulated visual constraints, \textit{Reflector} and \textit{Planner} remains static over time. To address this, we introduce a \textit{Memory} mechanism to support the evolution of their abilities.
Specifically, we maintain a \textit{Short Memory} $\mathcal{M}_s$ which stores step-level data within each episode of executing $l_i$ to support fine-grained updates for \textit{Reflector}:
\begin{equation}
\begin{aligned}
    &\text{Short Memory: } (o_t, o_{t-h}, l_i, p_t, c) \rightarrow \mathcal{M}_s\\
    &\text{Reflector w/ Memory: } c=F_r(\mathcal{M}_s)
\end{aligned}
\label{equ:short_mem}
\end{equation}
Here, instead of reflecting solely on the current transition $(o_t, o_{t-h}, p_t, l_i)$, the \textit{Reflector} can reason over the entire history of monitoring and reflection interactions stored in $\mathcal{M}_s$, allowing \textit{Reflector} to generate more accurate constraints by advancing Eq.~(\ref{equ:reflector}) with memory assistance as shown in Eq.~(\ref{equ:short_mem}).

For dynamical adaptation of \textit{Planner}, we introduce a \textit{Long Memory} module $\mathcal{M}_l$ to store high-level summaries for each episode involving instruction $l_i$. Accordingly, \textit{Planner} is updated from its original form in Eq.~(\ref{equ:planner}) to incorporate memory as shown in Eq.~(\ref{equ:long_mem}). To reduce storage cost, $\mathcal{M}_l$ logs only the initial and final frames $(o_{\rm init}, o_{\rm final})$ of the episode, the instruction $l_i$, and a textual summary $d_i$ of the VLA's execution behavior summarized from $\mathcal{M}_s$, i.e., $d_i = F_s(\mathcal{M}_s)$. This memory structure enables \textit{Planner} understand VLA proficiency on $l_{j\leq i}$ by analyzing both visual cues from visual transitions $(o_{\rm init}, o_{\rm final})$ and summarized behaviors $d_i$:
\begin{equation}
\begin{aligned}
&\text{Long Memory: } (o_{\rm init}, o_{\rm final}, l_i, d_i) \rightarrow \mathcal{M}_l\\
&\text{Planner w/ Memory: } l_{i+1} = F_p(o_t, l_{j\leq i}, l, M_l)
\end{aligned}
\label{equ:long_mem}
\end{equation}
Together, Short and Long Memories stores essential context for \textit{Reflector} and \textit{Planner}
to better understand both fine-grained VLA behaviors and high-level VLAs' capability boundary, enabling the agent frame dynamically adapt to VLAs according to their real-time performance.

\textbf{Marrying Agent Framework in Physical World via Embodied Toolbox}. 
To further close the gap toward physically grounded embodied agents, we introduce the Embodied Toolbox, a collection of tools that VLMs can query to support VLA execution and enhance overall system operation.

\begin{figure}[t]
\includegraphics[width=0.45\textwidth]{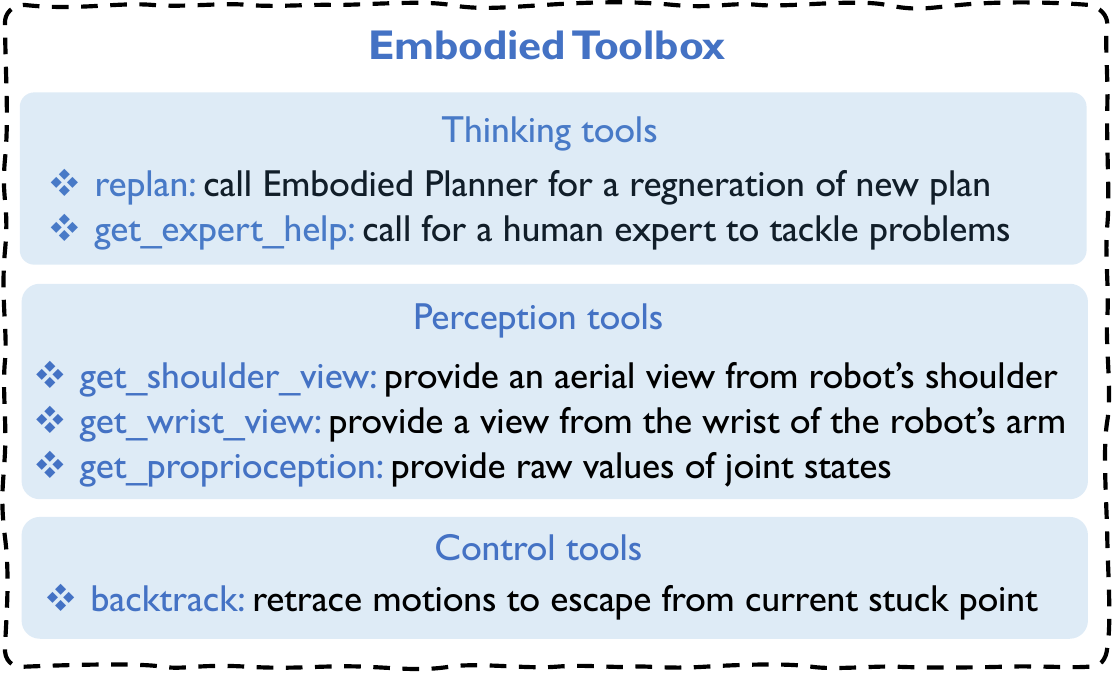}
\caption{An example of toolbox. It contains tools like perception, control and think to enhance the agent systems in real-world applications.}
\label{fig:toolbox}
\end{figure}

As shown in Figure~\ref{fig:toolbox}, the current toolbox includes three main categories: perception, reasoning, and control tools, with interaction examples illustrated in Figure~\ref{fig:example}.
The perception tool enables the agent to query multiple cameras, offering a broader perspective for improved scene understanding.
The reasoning tool assists in replanning or requesting human input when the agent encounters difficulties.
The control tool, such as backtrack, allows the system to reverse recent actions to recover from failure or local dead-ends—particularly useful when VLA performance is inconsistent and retries are beneficial.
Notice that the toolbox can be designed as extensible, allowing additional tools to be integrated through well-defined documentation.

\section{Experiments}
\label{sec:result}




\subsection{Experimental Setup}
\begin{figure}
\centering
\includegraphics[width=0.45\textwidth]{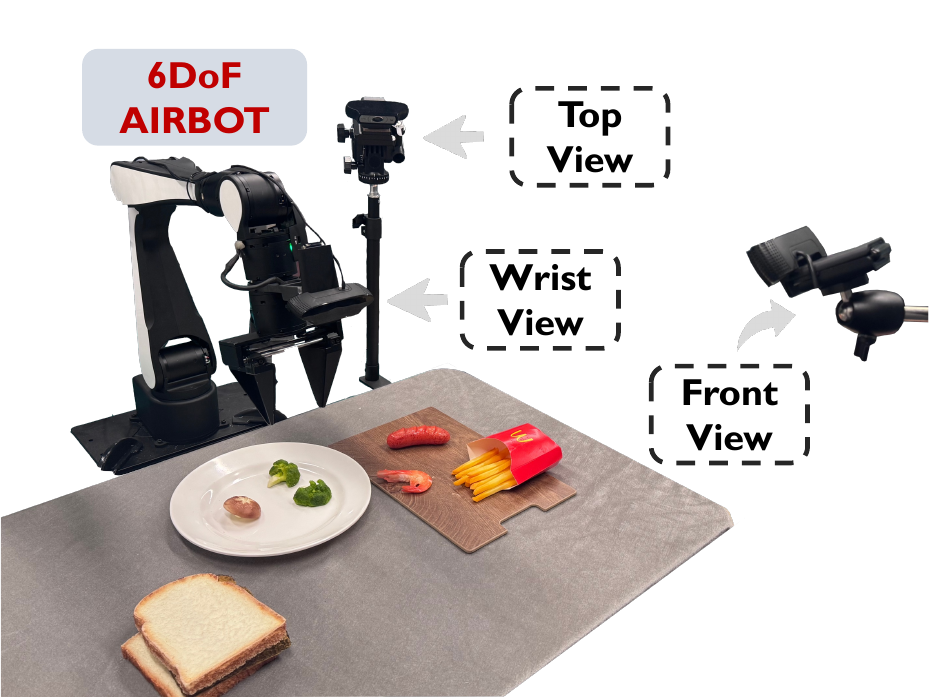}
\caption{The hardware setups for the real-world experiments.}

\vspace{-10pt}
\label{fig:real_setting}
\end{figure}

We conduct real-world experiments in a tabletop manipulation scenario. Specifically, we deploy AIRBOT, a 6-DOF robotic arm equipped with a gripper, as the primary platform for physical interaction. Three distinct RGB cameras are positioned to capture top-down, frontal, and wrist-mounted views of the workspace. The detailed hardware configuration is illustrated in Figure.~\ref{fig:real_setting}.

As an autonomous scaffolding, \name\ is compatible with arbitrary combinations of VLMs and VLAs, enabling coordinated interaction across components to fully leverage their respective capabilities. In this work, we deploy Gemini 2.0 Flash Lite as the \textit{Monitor}, and Gemini 2.0 Flash as both the \textit{Planner} and \textit{Reflector}. For the VLA component, we evaluate two instantiations of our framework: one using RDT-1B\cite{liu2024rdt} and the other using Diffusion Policy\cite{chi2024diffusionpolicy} as the low-level controller. All VLMs are used without any fine-tuning, and only prompt engineering is applied. The VLAs are fine-tuned on in-domain demonstrations to enhance their low-level manipulation capabilities. Further details on prompt construction and fine-tuning procedures are provided in Appendix~\ref{app:vla}.

\label{subsec:setup}


\textbf{Task Setup.} 
Our real-world benchmark comprises three distinct and challenging tabletop manipulation tasks, organized into two levels of complexity. The first level includes “Grab foods that contain dietary fiber” and “Grab foods that contain protein and fat”, while the second level features “Cook a meal”, a more demanding task that requires multi-step reasoning and execution. In all settings, the model is guided solely by high-level natural language instructions—such as “I’d like something meaty”—without access to any explicit low-level action commands. These abstract and ambiguous expressions require the system to infer user intent, reason over object semantics, and autonomously generate the appropriate manipulation sequence.

\textbf{Baselines.} 
We compare the performance of \name\ against two representative baselines: (1) a vanilla VLA policy, where control actions are directly predicted from observations and instructions without high-level reasoning, and (2) a conventional hierarchical framework, where a static VLM-based task planner generates subgoals to guide the policy. Since the static planner cannot autonomously track task progress and adjust subgoals in real time, we further consider a human-in-the-loop variant of this baseline. In this setting, a human operator monitors execution and manually prompts the VLM to regenerate subgoals as needed. This augmented version serves as a strong reference point, approximating the upper bound of performance under ideal high-level coordination.


\subsection{Main Results}
\label{subsec:main}

\begin{figure*}[t]
\centering
\includegraphics[width=1.0\textwidth]{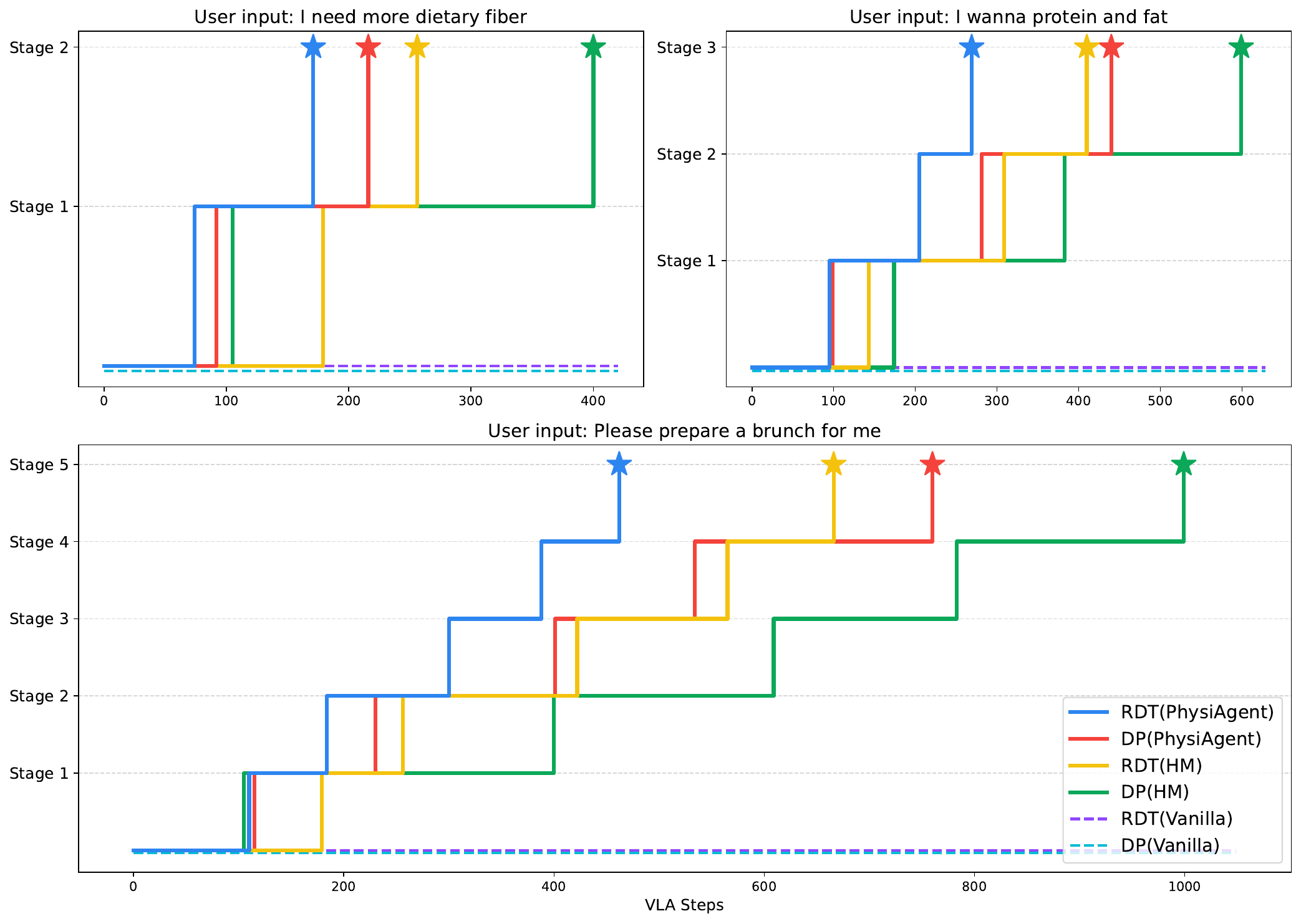}
\caption{Experimental results on tasks, such as ``I need more dietary fiber" and ``I want protein and fat", which involve no more than three VLA sub-tasks, and a task ``please prepare a brunch for me", which requires 5 stages. The Y-axis represents cumulative task progress, while the X-axis denotes VLA steps. Results show that our {PhysiAgent} framework consistently outperforms both vanilla and hierarchical baselines in terms of task completion efficiency.}
\label{fig:main_exp}
\end{figure*}

The main experimental results are presented in Figure.~\ref{fig:main_exp}, with performance averaged over five independent trials. The Y-axis represents cumulative task progress, while the X-axis denotes elapsed time (normalized by VLA steps). Each task—“Grab foods that contain dietary fiber,” “Grab foods that contain protein and fat,” and “Cook a meal”—is decomposed into 2, 3, and 5 discrete subtasks, respectively. Each subtask is defined by a rigorous checkpoint criterion, yet the agent retains full flexibility in execution order, selecting any available subtask at each timestep. Upon successful completion of a subtask, its corresponding stage is marked as completed and immediately contributes to the overall score, after which the remaining subtasks become available for pursuit.


As shown in Figure~\ref{fig:main_exp}, \name\ successfully completes nearly all stages across the three tasks with minimal execution steps, demonstrating both high efficiency. Notably, it extends the performance boundaries of both RDT and Diffusion Policy, enabling the execution of complex, unseen tasks and showing robustness even when operating with low-level policies. In contrast, vanilla VLA models struggle to interpret abstract language instructions and consistently fail to complete all tasks. Although human-in-the-loop hierarchical methods are also capable of task completion, their performance lags behind \name, likely due to insufficient interaction between the high-level planner and the low-level VLA. In comparison, \name\ achieves organic integration between VLMs and VLAs. Its unified architecture—comprising the \textit{Planner}, \textit{Reflector}, \textit{Monitor}, \textit{Toolbox}, and \textit{Memory}—forms a more intelligent, adaptive, and effective embodied system for real-world deployment.

\section{Related Works}
\label{sec:related_works}

\textbf{Visual Language Action Models:} Visual Language Action models have emerged as a cornerstone in the field of embodied artificial intelligence~\cite{ma2024survey,brohan2023rt,liu2024aligning}, demonstrating impressive capabilities across a range of tasks, from tabletop manipulation tasks~\cite{kim2024openvla,octo_2023,chi2024diffusionpolicy,liu2024rdt} to complex dexterous operations~\cite{zhong2025dexgraspvla,chen2025conrft}.
Benefiting from end-to-end training paradigms, VLA models offer a promising path toward scalable embodied learning. Recent efforts in data collection and synthetic data generation~\cite{shi2025hi,maddukuri2025sim} have further accelerated the development of generalized, foundation-level VLA models. However, despite continued scaling, recent studies~\cite{blank2024scaling} have reported diminishing returns in performance as data volume increases, suggesting that naively increasing data is not an efficient path forward.

\textbf{Hierarchical Framework for Embodied AI:}
A straightforward approach to enhance the generalization of embodied model is to integrate VLA models with powerful Vision-Language Models (VLMs).
One line of work focuses on joint training of VLMs and low-level action models. These methods treat the VLM as an embedded component within an encoder–decoder-style VLA pipeline~\cite{bjorck2025gr00t, bu2024towards, lee2024behavior, zheng2025universal, szot2024grounding}, leveraging pretrained parameters for efficient knowledge transfer. Another line of research uses natural language as an interface to connect VLMs and VLA models~\cite{beyer2024paligemma}. For instance, Ahn et al.\cite{ahn2022can} employ an LLM to select skill-level instructions, guided by a pretrained value function. Huang et al.\cite{huang2022inner} further extend this paradigm by incorporating real-time environmental feedback via natural language. While these approaches highlight the potential of integrating VLMs and VLA models, they often rely on rigid combination mechanism or task-specific training—leaving open the question of how to achieve more flexible, autonomous, and modular integration between VLMs and VLAs.


\textbf{LLM-based Agents in Digital World:} LLM-based agents have recently gained increasing attention for their ability to perform complex, instruction-driven behaviors in simulated environments. Much of this work focuses on single-agent settings, where LLMs are used to reason over task structures and synthesize grounded, skill-level actions. For example, Huang et al.\cite{huang2022language} apply zero-shot prompting for instruction following in VirtualHome\cite{puig2018virtualhome}, while LLM-Planner~\cite{song2023llm} leverages few-shot prompting to support long-horizon planning in ALFRED~\cite{shridhar2020alfred}. With the advancement of interactive simulators such as MineDojo~\cite{fan2022minedojo} and MineRL~\cite{guss2019minerl}, the grounded capabilities of LLM-based agents have been further explored in more complex, open-ended environments~\cite{wang2023voyager, qin2024mp5}. More recent efforts have extended this paradigm to multi-agent systems, where multiple specialized agents collaborate to solve coordinated tasks~\cite{li2023camel}. While these approaches demonstrate the promise of LLM-based agents in simulation, they remain confined to virtual domains and largely overlook the unique challenges of real-world deployment—such as egocentric perception, asynchronous execution, and physically grounded interaction.








\section{Conclusion}
\label{sec:conclusion}
We present \name, the first physically grounded embodied agent framework that enables dynamic and adaptive collaboration between VLMs and VLAs through a unified, self-regulating architecture. By leveraging reflection, monitoring, memory, and an extensible toolbox, \name~goes beyond rigid pipelines to support real-time feedback from task execution. Our experiments in a real-world robotic platform validates the framework’s effectiveness, demonstrating robust generalization across tasks. Importantly, \name~is modular and plug-and-play, making it readily applicable to a wide range of VLMs and VLAs with minimal engineering overhead. We believe \name~takes a crucial step toward building scalable, general-purpose embodied agents capable of operating autonomously in complex physical environments.

\section*{Acknowledgement}
\label{Acknowledgements}
This work is supported by funding from Wuxi Research Institute of Applied Technologies, Tsinghua University under Grant 20242001120.

\section*{Impact Statement}

Our PhysiAgent framework aims to integrate VLMs and VLAs, effectively grounding embodied agent frameworks in real-world settings. There are many potential societal consequences of our work, none of which we feel must be specifically highlighted here.

\bibliography{example_paper}

\begin{thebibliography}{54}
\providecommand{\natexlab}[1]{#1}
\providecommand{\url}[1]{\texttt{#1}}
\expandafter\ifx\csname urlstyle\endcsname\relax
  \providecommand{\doi}[1]{doi: #1}\else
  \providecommand{\doi}{doi: \begingroup \urlstyle{rm}\Url}\fi

\bibitem[Achiam et~al.(2023)Achiam, Adler, Agarwal, Ahmad, Akkaya, Aleman, Almeida, Altenschmidt, Altman, Anadkat, et~al.]{achiam2023gpt}
Achiam, J., Adler, S., Agarwal, S., Ahmad, L., Akkaya, I., Aleman, F.~L., Almeida, D., Altenschmidt, J., Altman, S., Anadkat, S., et~al.
\newblock Gpt-4 technical report.
\newblock \emph{arXiv preprint arXiv:2303.08774}, 2023.

\bibitem[Ahn et~al.(2022)Ahn, Brohan, Brown, Chebotar, Cortes, David, Finn, Fu, Gopalakrishnan, Hausman, et~al.]{ahn2022can}
Ahn, M., Brohan, A., Brown, N., Chebotar, Y., Cortes, O., David, B., Finn, C., Fu, C., Gopalakrishnan, K., Hausman, K., et~al.
\newblock Do as i can, not as i say: Grounding language in robotic affordances.
\newblock \emph{arXiv preprint arXiv:2204.01691}, 2022.

\bibitem[Beyer et~al.(2024)Beyer, Steiner, Pinto, Kolesnikov, Wang, Salz, Neumann, Alabdulmohsin, Tschannen, Bugliarello, et~al.]{beyer2024paligemma}
Beyer, L., Steiner, A., Pinto, A.~S., Kolesnikov, A., Wang, X., Salz, D., Neumann, M., Alabdulmohsin, I., Tschannen, M., Bugliarello, E., et~al.
\newblock Paligemma: A versatile 3b vlm for transfer.
\newblock \emph{arXiv preprint arXiv:2407.07726}, 2024.

\bibitem[Bjorck et~al.(2025)Bjorck, Casta{\~n}eda, Cherniadev, Da, Ding, Fan, Fang, Fox, Hu, Huang, et~al.]{bjorck2025gr00t}
Bjorck, J., Casta{\~n}eda, F., Cherniadev, N., Da, X., Ding, R., Fan, L., Fang, Y., Fox, D., Hu, F., Huang, S., et~al.
\newblock Gr00t n1: An open foundation model for generalist humanoid robots.
\newblock \emph{arXiv preprint arXiv:2503.14734}, 2025.

\bibitem[Black et~al.(2024)Black, Brown, Driess, Esmail, Equi, Finn, Fusai, Groom, Hausman, Ichter, et~al.]{black2024pi_0}
Black, K., Brown, N., Driess, D., Esmail, A., Equi, M., Finn, C., Fusai, N., Groom, L., Hausman, K., Ichter, B., et~al.
\newblock $\pi_0 $: A vision-language-action flow model for general robot control.
\newblock \emph{arXiv preprint arXiv:2410.24164}, 2024.

\bibitem[Blank et~al.(2024)Blank, Reuss, R{\"u}hle, Ya{\u{g}}murlu, Wenzel, Mees, and Lioutikov]{blank2024scaling}
Blank, N., Reuss, M., R{\"u}hle, M., Ya{\u{g}}murlu, {\"O}.~E., Wenzel, F., Mees, O., and Lioutikov, R.
\newblock Scaling robot policy learning via zero-shot labeling with foundation models.
\newblock \emph{arXiv preprint arXiv:2410.17772}, 2024.

\bibitem[Brohan et~al.(2023)Brohan, Brown, Carbajal, Chebotar, Chen, Choromanski, Ding, Driess, Dubey, Finn, et~al.]{brohan2023rt}
Brohan, A., Brown, N., Carbajal, J., Chebotar, Y., Chen, X., Choromanski, K., Ding, T., Driess, D., Dubey, A., Finn, C., et~al.
\newblock Rt-2: Vision-language-action models transfer web knowledge to robotic control.
\newblock \emph{arXiv preprint arXiv:2307.15818}, 2023.

\bibitem[Bruce et~al.(2024)Bruce, Dennis, Edwards, Parker-Holder, Shi, Hughes, Lai, Mavalankar, Steigerwald, Apps, et~al.]{bruce2024genie}
Bruce, J., Dennis, M.~D., Edwards, A., Parker-Holder, J., Shi, Y., Hughes, E., Lai, M., Mavalankar, A., Steigerwald, R., Apps, C., et~al.
\newblock Genie: Generative interactive environments.
\newblock In \emph{International Conference on Machine Learning}, pp.\  4603--4623. PMLR, 2024.

\bibitem[Bu et~al.(2024)Bu, Li, Chen, Cai, Zeng, Cui, Yao, and Qiao]{bu2024towards}
Bu, Q., Li, H., Chen, L., Cai, J., Zeng, J., Cui, H., Yao, M., and Qiao, Y.
\newblock Towards synergistic, generalized, and efficient dual-system for robotic manipulation.
\newblock \emph{arXiv preprint arXiv:2410.08001}, 2024.

\bibitem[Chen et~al.(2024)Chen, Xu, Dharmarajan, Irshad, Cheng, Keutzer, Tomizuka, Vuong, and Goldberg]{chen2024roviaug}
Chen, L.~Y., Xu, C., Dharmarajan, K., Irshad, M.~Z., Cheng, R., Keutzer, K., Tomizuka, M., Vuong, Q., and Goldberg, K.
\newblock Rovi-aug: Robot and viewpoint augmentation for cross-embodiment robot learning.
\newblock In \emph{Conference on Robot Learning (CoRL)}, Munich, Germany, 2024.

\bibitem[Chen et~al.(2025)Chen, Tian, Zhou, Liu, Li, and Zhao]{chen2025conrft}
Chen, Y., Tian, S., Zhou, Y., Liu, S., Li, H., and Zhao, D.
\newblock Conrft: A reinforced fine-tuning method for vla models via consistency policy.
\newblock \emph{arXiv preprint arXiv:2502.05450}, 2025.

\bibitem[Chi et~al.(2024)Chi, Xu, Feng, Cousineau, Du, Burchfiel, Tedrake, and Song]{chi2024diffusionpolicy}
Chi, C., Xu, Z., Feng, S., Cousineau, E., Du, Y., Burchfiel, B., Tedrake, R., and Song, S.
\newblock Diffusion policy: Visuomotor policy learning via action diffusion.
\newblock \emph{The International Journal of Robotics Research}, 2024.

\bibitem[DeepMind(2024)]{gemini-2-flash}
DeepMind, G.
\newblock Gemini 2.0 flash thinking.
\newblock Technical report, https://deepmind.google/technologies/ gemini/flash-thinking, 2024.

\bibitem[Fan et~al.(2022)Fan, Wang, Jiang, Mandlekar, Yang, Zhu, Tang, Huang, Zhu, and Anandkumar]{fan2022minedojo}
Fan, L., Wang, G., Jiang, Y., Mandlekar, A., Yang, Y., Zhu, H., Tang, A., Huang, D.-A., Zhu, Y., and Anandkumar, A.
\newblock Minedojo: Building open-ended embodied agents with internet-scale knowledge.
\newblock \emph{Advances in Neural Information Processing Systems}, 35:\penalty0 18343--18362, 2022.

\bibitem[Guss et~al.(2019)Guss, Houghton, Topin, Wang, Codel, Veloso, and Salakhutdinov]{guss2019minerl}
Guss, W.~H., Houghton, B., Topin, N., Wang, P., Codel, C., Veloso, M., and Salakhutdinov, R.
\newblock Minerl: A large-scale dataset of minecraft demonstrations.
\newblock \emph{arXiv preprint arXiv:1907.13440}, 2019.

\bibitem[Huang et~al.(2022{\natexlab{a}})Huang, Abbeel, Pathak, and Mordatch]{huang2022language}
Huang, W., Abbeel, P., Pathak, D., and Mordatch, I.
\newblock Language models as zero-shot planners: Extracting actionable knowledge for embodied agents.
\newblock In \emph{International conference on machine learning}, pp.\  9118--9147. PMLR, 2022{\natexlab{a}}.

\bibitem[Huang et~al.(2022{\natexlab{b}})Huang, Xia, Xiao, Chan, Liang, Florence, Zeng, Tompson, Mordatch, Chebotar, et~al.]{huang2022inner}
Huang, W., Xia, F., Xiao, T., Chan, H., Liang, J., Florence, P., Zeng, A., Tompson, J., Mordatch, I., Chebotar, Y., et~al.
\newblock Inner monologue: Embodied reasoning through planning with language models.
\newblock \emph{arXiv preprint arXiv:2207.05608}, 2022{\natexlab{b}}.

\bibitem[Hurst et~al.(2024)Hurst, Lerer, Goucher, Perelman, Ramesh, Clark, Ostrow, Welihinda, Hayes, Radford, et~al.]{hurst2024gpt}
Hurst, A., Lerer, A., Goucher, A.~P., Perelman, A., Ramesh, A., Clark, A., Ostrow, A., Welihinda, A., Hayes, A., Radford, A., et~al.
\newblock Gpt-4o system card.
\newblock \emph{arXiv preprint arXiv:2410.21276}, 2024.

\bibitem[Intelligence et~al.(2025)Intelligence, Black, Brown, Darpinian, Dhabalia, Driess, Esmail, Equi, Finn, Fusai, Galliker, Ghosh, Groom, Hausman, Ichter, Jakubczak, Jones, Ke, LeBlanc, Levine, Li-Bell, Mothukuri, Nair, Pertsch, Ren, Shi, Smith, Springenberg, Stachowicz, Tanner, Vuong, Walke, Walling, Wang, Yu, and Zhilinsky]{intelligence2025pi05visionlanguageactionmodelopenworld}
Intelligence, P., Black, K., Brown, N., Darpinian, J., Dhabalia, K., Driess, D., Esmail, A., Equi, M., Finn, C., Fusai, N., Galliker, M.~Y., Ghosh, D., Groom, L., Hausman, K., Ichter, B., Jakubczak, S., Jones, T., Ke, L., LeBlanc, D., Levine, S., Li-Bell, A., Mothukuri, M., Nair, S., Pertsch, K., Ren, A.~Z., Shi, L.~X., Smith, L., Springenberg, J.~T., Stachowicz, K., Tanner, J., Vuong, Q., Walke, H., Walling, A., Wang, H., Yu, L., and Zhilinsky, U.
\newblock $\pi_{0.5}$: a vision-language-action model with open-world generalization, 2025.
\newblock URL \url{https://arxiv.org/abs/2504.16054}.

\bibitem[Kim et~al.(2024)Kim, Pertsch, Karamcheti, Xiao, Balakrishna, Nair, Rafailov, Foster, Lam, Sanketi, et~al.]{kim2024openvla}
Kim, M.~J., Pertsch, K., Karamcheti, S., Xiao, T., Balakrishna, A., Nair, S., Rafailov, R., Foster, E., Lam, G., Sanketi, P., et~al.
\newblock Openvla: An open-source vision-language-action model.
\newblock \emph{arXiv preprint arXiv:2406.09246}, 2024.

\bibitem[Lee et~al.(2024)Lee, Wang, Etukuru, Kim, Shafiullah, and Pinto]{lee2024behavior}
Lee, S., Wang, Y., Etukuru, H., Kim, H.~J., Shafiullah, N. M.~M., and Pinto, L.
\newblock Behavior generation with latent actions.
\newblock \emph{arXiv preprint arXiv:2403.03181}, 2024.

\bibitem[Li et~al.(2023)Li, Hammoud, Itani, Khizbullin, and Ghanem]{li2023camel}
Li, G., Hammoud, H., Itani, H., Khizbullin, D., and Ghanem, B.
\newblock Camel: Communicative agents for" mind" exploration of large language model society.
\newblock \emph{Advances in Neural Information Processing Systems}, 36:\penalty0 51991--52008, 2023.

\bibitem[Li et~al.(2024{\natexlab{a}})Li, Lai, Li, Ren, Zhang, Kang, Wang, Li, Zhang, Ma, et~al.]{li2024agent}
Li, J., Lai, Y., Li, W., Ren, J., Zhang, M., Kang, X., Wang, S., Li, P., Zhang, Y.-Q., Ma, W., et~al.
\newblock Agent hospital: A simulacrum of hospital with evolvable medical agents.
\newblock \emph{arXiv preprint arXiv:2405.02957}, 2024{\natexlab{a}}.

\bibitem[Li et~al.(2024{\natexlab{b}})Li, Zheng, Zheng, Mao, Hu, Cheng, Niu, Liu, Liu, Liu, et~al.]{li2024decisionnce}
Li, J., Zheng, J., Zheng, Y., Mao, L., Hu, X., Cheng, S., Niu, H., Liu, J., Liu, Y., Liu, J., et~al.
\newblock Decisionnce: Embodied multimodal representations via implicit preference learning.
\newblock In \emph{International Conference on Machine Learning}, pp.\  29461--29488. PMLR, 2024{\natexlab{b}}.

\bibitem[Lin et~al.(2024)Lin, Hu, Sheng, Wen, You, and Gao]{lin2024data}
Lin, F., Hu, Y., Sheng, P., Wen, C., You, J., and Gao, Y.
\newblock Data scaling laws in imitation learning for robotic manipulation.
\newblock \emph{arXiv preprint arXiv:2410.18647}, 2024.

\bibitem[Liu et~al.(2024{\natexlab{a}})Liu, Wu, Li, Tan, Chen, Wang, Xu, Su, and Zhu]{liu2024rdt}
Liu, S., Wu, L., Li, B., Tan, H., Chen, H., Wang, Z., Xu, K., Su, H., and Zhu, J.
\newblock Rdt-1b: a diffusion foundation model for bimanual manipulation.
\newblock \emph{arXiv preprint arXiv:2410.07864}, 2024{\natexlab{a}}.

\bibitem[Liu et~al.(2024{\natexlab{b}})Liu, Chen, Bai, Liang, Li, Gao, and Lin]{liu2024aligning}
Liu, Y., Chen, W., Bai, Y., Liang, X., Li, G., Gao, W., and Lin, L.
\newblock Aligning cyber space with physical world: A comprehensive survey on embodied ai.
\newblock \emph{arXiv preprint arXiv:2407.06886}, 2024{\natexlab{b}}.

\bibitem[Ma et~al.(2024)Ma, Song, Zhuang, Hao, and King]{ma2024survey}
Ma, Y., Song, Z., Zhuang, Y., Hao, J., and King, I.
\newblock A survey on vision-language-action models for embodied ai.
\newblock \emph{arXiv preprint arXiv:2405.14093}, 2024.

\bibitem[Ma et~al.(2025)Ma, Hejna, Fu, Shah, Liang, Xu, Kirmani, Xu, Driess, Xiao, Bastani, Jayaraman, Yu, Zhang, Sadigh, and Xia]{ma2025vision}
Ma, Y.~J., Hejna, J., Fu, C., Shah, D., Liang, J., Xu, Z., Kirmani, S., Xu, P., Driess, D., Xiao, T., Bastani, O., Jayaraman, D., Yu, W., Zhang, T., Sadigh, D., and Xia, F.
\newblock Vision language models are in-context value learners.
\newblock In \emph{The Thirteenth International Conference on Learning Representations}, 2025.
\newblock URL \url{https://openreview.net/forum?id=friHAl5ofG}.

\bibitem[Maddukuri et~al.(2025)Maddukuri, Jiang, Chen, Nasiriany, Xie, Fang, Huang, Wang, Xu, Chernyadev, et~al.]{maddukuri2025sim}
Maddukuri, A., Jiang, Z., Chen, L.~Y., Nasiriany, S., Xie, Y., Fang, Y., Huang, W., Wang, Z., Xu, Z., Chernyadev, N., et~al.
\newblock Sim-and-real co-training: A simple recipe for vision-based robotic manipulation.
\newblock \emph{arXiv preprint arXiv:2503.24361}, 2025.

\bibitem[Nguyen et~al.(2024)Nguyen, Chen, Wang, Wu, Park, Hu, Lyu, Wu, Aponte, Xia, et~al.]{nguyen2024gui}
Nguyen, D., Chen, J., Wang, Y., Wu, G., Park, N., Hu, Z., Lyu, H., Wu, J., Aponte, R., Xia, Y., et~al.
\newblock Gui agents: A survey.
\newblock \emph{arXiv preprint arXiv:2412.13501}, 2024.

\bibitem[{Octo Model Team} et~al.(2024){Octo Model Team}, Ghosh, Walke, Pertsch, Black, Mees, Dasari, Hejna, Xu, Luo, Kreiman, Tan, Chen, Sanketi, Vuong, Xiao, Sadigh, Finn, and Levine]{octo_2023}
{Octo Model Team}, Ghosh, D., Walke, H., Pertsch, K., Black, K., Mees, O., Dasari, S., Hejna, J., Xu, C., Luo, J., Kreiman, T., Tan, Y., Chen, L.~Y., Sanketi, P., Vuong, Q., Xiao, T., Sadigh, D., Finn, C., and Levine, S.
\newblock Octo: An open-source generalist robot policy.
\newblock In \emph{Proceedings of Robotics: Science and Systems}, Delft, Netherlands, 2024.

\bibitem[Park et~al.(2023)Park, O'Brien, Cai, Morris, Liang, and Bernstein]{park2023generative}
Park, J.~S., O'Brien, J., Cai, C.~J., Morris, M.~R., Liang, P., and Bernstein, M.~S.
\newblock Generative agents: Interactive simulacra of human behavior.
\newblock In \emph{Proceedings of the 36th annual acm symposium on user interface software and technology}, pp.\  1--22, 2023.

\bibitem[Puig et~al.(2018)Puig, Ra, Boben, Li, Wang, Fidler, and Torralba]{puig2018virtualhome}
Puig, X., Ra, K., Boben, M., Li, J., Wang, T., Fidler, S., and Torralba, A.
\newblock Virtualhome: Simulating household activities via programs.
\newblock In \emph{Proceedings of the IEEE conference on computer vision and pattern recognition}, pp.\  8494--8502, 2018.

\bibitem[Qin et~al.(2024)Qin, Zhou, Liu, Yin, Sheng, Zhang, Qiao, and Shao]{qin2024mp5}
Qin, Y., Zhou, E., Liu, Q., Yin, Z., Sheng, L., Zhang, R., Qiao, Y., and Shao, J.
\newblock Mp5: A multi-modal open-ended embodied system in minecraft via active perception.
\newblock In \emph{2024 IEEE/CVF Conference on Computer Vision and Pattern Recognition (CVPR)}, pp.\  16307--16316. IEEE, 2024.

\bibitem[Reuss et~al.(2025)Reuss, Zhou, R{\"u}hle, Ya{\u{g}}murlu, Otto, and Lioutikov]{reussflower}
Reuss, M., Zhou, H., R{\"u}hle, M., Ya{\u{g}}murlu, {\"O}.~E., Otto, F., and Lioutikov, R.
\newblock Flower: Democratizing generalist robot policies with efficient vision-language-action flow policies.
\newblock In \emph{7th Robot Learning Workshop: Towards Robots with Human-Level Abilities}, 2025.

\bibitem[Shi et~al.(2024)Shi, Hu, Zhao, Sharma, Pertsch, Luo, Levine, and Finn]{shi2024yell}
Shi, L.~X., Hu, Z., Zhao, T.~Z., Sharma, A., Pertsch, K., Luo, J., Levine, S., and Finn, C.
\newblock Yell at your robot: Improving on-the-fly from language corrections.
\newblock \emph{arXiv preprint arXiv:2403.12910}, 2024.

\bibitem[Shi et~al.(2025)Shi, Ichter, Equi, Ke, Pertsch, Vuong, Tanner, Walling, Wang, Fusai, et~al.]{shi2025hi}
Shi, L.~X., Ichter, B., Equi, M., Ke, L., Pertsch, K., Vuong, Q., Tanner, J., Walling, A., Wang, H., Fusai, N., et~al.
\newblock Hi robot: Open-ended instruction following with hierarchical vision-language-action models.
\newblock \emph{arXiv preprint arXiv:2502.19417}, 2025.

\bibitem[Shridhar et~al.(2020)Shridhar, Thomason, Gordon, Bisk, Han, Mottaghi, Zettlemoyer, and Fox]{shridhar2020alfred}
Shridhar, M., Thomason, J., Gordon, D., Bisk, Y., Han, W., Mottaghi, R., Zettlemoyer, L., and Fox, D.
\newblock Alfred: A benchmark for interpreting grounded instructions for everyday tasks.
\newblock In \emph{Proceedings of the IEEE/CVF conference on computer vision and pattern recognition}, pp.\  10740--10749, 2020.

\bibitem[Song et~al.(2023)Song, Wu, Washington, Sadler, Chao, and Su]{song2023llm}
Song, C.~H., Wu, J., Washington, C., Sadler, B.~M., Chao, W.-L., and Su, Y.
\newblock Llm-planner: Few-shot grounded planning for embodied agents with large language models.
\newblock In \emph{Proceedings of the IEEE/CVF international conference on computer vision}, pp.\  2998--3009, 2023.

\bibitem[Szot et~al.(2024)Szot, Mazoure, Agrawal, Hjelm, Kira, and Toshev]{szot2024grounding}
Szot, A., Mazoure, B., Agrawal, H., Hjelm, R.~D., Kira, Z., and Toshev, A.
\newblock Grounding multimodal large language models in actions.
\newblock \emph{Advances in Neural Information Processing Systems}, 37:\penalty0 20198--20224, 2024.

\bibitem[Team(2025)]{doubao-v}
Team, D.
\newblock Doubao-vision-pro-32k.
\newblock Technical report, https://www.volcengine.com/product/doubao, 2025.

\bibitem[Wang et~al.(2023{\natexlab{a}})Wang, Xie, Jiang, Mandlekar, Xiao, Zhu, Fan, and Anandkumar]{wang2023voyager}
Wang, G., Xie, Y., Jiang, Y., Mandlekar, A., Xiao, C., Zhu, Y., Fan, L., and Anandkumar, A.
\newblock Voyager: An open-ended embodied agent with large language models.
\newblock \emph{arXiv preprint arXiv:2305.16291}, 2023{\natexlab{a}}.

\bibitem[Wang et~al.(2024{\natexlab{a}})Wang, Ma, Feng, Zhang, Yang, Zhang, Chen, Tang, Chen, Lin, et~al.]{wang2024survey}
Wang, L., Ma, C., Feng, X., Zhang, Z., Yang, H., Zhang, J., Chen, Z., Tang, J., Chen, X., Lin, Y., et~al.
\newblock A survey on large language model based autonomous agents.
\newblock \emph{Frontiers of Computer Science}, 18\penalty0 (6):\penalty0 186345, 2024{\natexlab{a}}.

\bibitem[Wang et~al.(2023{\natexlab{b}})Wang, Cai, Chen, Liu, Ma, and Liang]{wang2023describe}
Wang, Z., Cai, S., Chen, G., Liu, A., Ma, X., and Liang, Y.
\newblock Describe, explain, plan and select: Interactive planning with large language models enables open-world multi-task agents.
\newblock \emph{arXiv preprint arXiv:2302.01560}, 2023{\natexlab{b}}.

\bibitem[Wang et~al.(2024{\natexlab{b}})Wang, Cai, Liu, Jin, Hou, Zhang, Lin, He, Zheng, Yang, et~al.]{wang2024jarvis}
Wang, Z., Cai, S., Liu, A., Jin, Y., Hou, J., Zhang, B., Lin, H., He, Z., Zheng, Z., Yang, Y., et~al.
\newblock Jarvis-1: Open-world multi-task agents with memory-augmented multimodal language models.
\newblock \emph{IEEE Transactions on Pattern Analysis and Machine Intelligence}, 2024{\natexlab{b}}.

\bibitem[Xi et~al.(2025)Xi, Chen, Guo, He, Ding, Hong, Zhang, Wang, Jin, Zhou, et~al.]{xi2025rise}
Xi, Z., Chen, W., Guo, X., He, W., Ding, Y., Hong, B., Zhang, M., Wang, J., Jin, S., Zhou, E., et~al.
\newblock The rise and potential of large language model based agents: A survey.
\newblock \emph{Science China Information Sciences}, 68\penalty0 (2):\penalty0 121101, 2025.

\bibitem[Yang et~al.(2024)Yang, Zhang, Zheng, Jiang, Gan, Wang, Ling, Chen, Ma, Dong, et~al.]{yang2024oasis}
Yang, Z., Zhang, Z., Zheng, Z., Jiang, Y., Gan, Z., Wang, Z., Ling, Z., Chen, J., Ma, M., Dong, B., et~al.
\newblock Oasis: Open agents social interaction simulations on one million agents.
\newblock \emph{arXiv preprint arXiv:2411.11581}, 2024.

\bibitem[Ye et~al.(2025)Ye, Jang, Jeon, Joo, Yang, Peng, Mandlekar, Tan, Chao, Lin, Liden, Lee, Gao, Zettlemoyer, Fox, and Seo]{ye2025latent}
Ye, S., Jang, J., Jeon, B., Joo, S.~J., Yang, J., Peng, B., Mandlekar, A., Tan, R., Chao, Y.-W., Lin, B.~Y., Liden, L., Lee, K., Gao, J., Zettlemoyer, L., Fox, D., and Seo, M.
\newblock Latent action pretraining from videos.
\newblock In \emph{The Thirteenth International Conference on Learning Representations}, 2025.
\newblock URL \url{https://openreview.net/forum?id=VYOe2eBQeh}.

\bibitem[Zhang et~al.(2024{\natexlab{a}})Zhang, He, Qian, Li, Li, Qin, Kang, Ma, Liu, Lin, et~al.]{zhang2024large}
Zhang, C., He, S., Qian, J., Li, B., Li, L., Qin, S., Kang, Y., Ma, M., Liu, G., Lin, Q., et~al.
\newblock Large language model-brained gui agents: A survey.
\newblock \emph{arXiv preprint arXiv:2411.18279}, 2024{\natexlab{a}}.

\bibitem[Zhang et~al.(2024{\natexlab{b}})Zhang, Wang, Wang, Li, Liu, Wei, Wang, Zhang, and Wang]{zhang2024uni}
Zhang, J., Wang, K., Wang, S., Li, M., Liu, H., Wei, S., Wang, Z., Zhang, Z., and Wang, H.
\newblock Uni-navid: A video-based vision-language-action model for unifying embodied navigation tasks.
\newblock \emph{arXiv preprint arXiv:2412.06224}, 2024{\natexlab{b}}.

\bibitem[Zhang et~al.(2024{\natexlab{c}})Zhang, Wang, Xu, Zhou, Hong, Fang, Wu, Zhang, and Wang]{zhang2024navid}
Zhang, J., Wang, K., Xu, R., Zhou, G., Hong, Y., Fang, X., Wu, Q., Zhang, Z., and Wang, H.
\newblock Navid: Video-based vlm plans the next step for vision-and-language navigation.
\newblock \emph{arXiv preprint arXiv:2402.15852}, 2024{\natexlab{c}}.

\bibitem[Zheng et~al.(2025)Zheng, Li, Liu, Zheng, Wang, Ou, Liu, Liu, Zhang, and Zhan]{zheng2025universal}
Zheng, J., Li, J., Liu, D., Zheng, Y., Wang, Z., Ou, Z., Liu, Y., Liu, J., Zhang, Y.-Q., and Zhan, X.
\newblock Universal actions for enhanced embodied foundation models.
\newblock \emph{arXiv preprint arXiv:2501.10105}, 2025.

\bibitem[Zhong et~al.(2025)Zhong, Huang, Li, Zhang, Liang, Yang, and Chen]{zhong2025dexgraspvla}
Zhong, Y., Huang, X., Li, R., Zhang, C., Liang, Y., Yang, Y., and Chen, Y.
\newblock Dexgraspvla: A vision-language-action framework towards general dexterous grasping.
\newblock \emph{arXiv preprint arXiv:2502.20900}, 2025.

\end{thebibliography}
\bibliographystyle{icml2025}

\newpage
\appendix
\onecolumn

\section{Limitations}
\label{sec:limitations}

Although the effectiveness of \name~has been demonstrated in real-world robotic tabletop manipulation scenarios, its applicability to other categories of embodied AI tasks—such as navigation~\cite{zhang2024uni}—remains unexplored in the physical world. Given that \name~is designed with modularity and task-type agnosticism in mind, future work may extend its deployment to a broader range of embodied AI domains under real-world settings.

Furthermore, the current instantiation of \name~focuses on enabling full autonomy within a single-agent embodied framework. Other critical components of a holistic embodied agent system—such as inter-agent communication and collaboration—have yet to be integrated. Incorporating communication modules, as suggested in~\cite{park2023generative}, could facilitate the development of multi-agent systems, thereby advancing the capabilities of embodied AI frameworks.

Lastly, our implementation relies exclusively on commercially available vision-language models (VLMs), including Gemini 2.0 Flash~\cite{gemini-2-flash}, GPT-4o~\cite{hurst2024gpt}, and Doubao-Vision-Pro-32k~\cite{doubao-v}. Replacing these proprietary models with open-source alternatives holds promise for constructing more transparent, customizable, and fully embedded embodied systems.

\section{Detailed Workflow Illustration}
\label{app:workflow}

\begin{figure*}[t]
\centering
\includegraphics[width=1.00\textwidth]{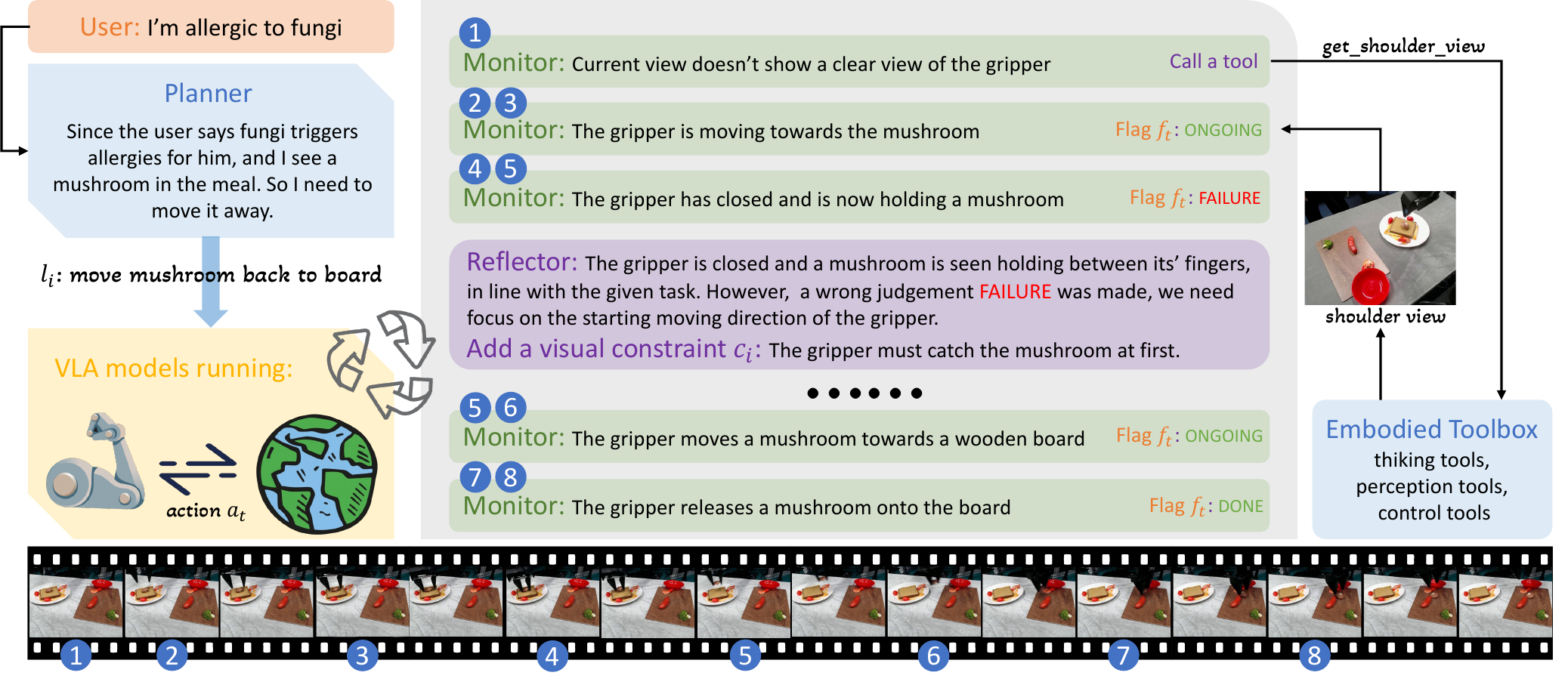}
\vspace{-20pt}
\caption{\small A detailed illustration of the PhysiAgent workflow for a single-stage task.}
\label{fig:single_example}
\vspace{-10pt}
\end{figure*}

Figure~\ref{fig:single_example} provides a detailed visualization of the \name~workflow when executing a single-stage task. This figure serves as an expanded illustration of the high-level overview presented in Figure~\ref{fig:example}.

\section{Real-world Tabletop Manipulation Dataset}
\label{app:dataset}

The VLA tasks used in our primary experiments include: \textit{put broccoli on plate}, \textit{put mushroom on plate}, \textit{put sausage on plate}, \textit{put shrimp on plate}, and \textit{put chips on plate}. Each task comprises 150 human teleoperated demonstrations, collected under real-world tabletop manipulation settings.

\section{Details of Implemented Vision-Language-Action Models}
\label{app:vla}

We implement a multi-task variant of the Diffusion Policy, employing a ResNet-50 as the visual backbone. Language instructions are incorporated via FiLM conditioning layers. This model is trained on the dataset described in Appendix~\ref{app:dataset} for 1.2 million steps using 4 NVIDIA A800 GPUs in 27 hours, with a batch size of 64 and a learning rate of 0.0003.

In addition, we fine-tune the RDT-1B model on our domain-specific dataset for 50,000 steps using 8 A800 GPUs in 20 hours, maintaining a batch size of 64 and a learning rate of 0.0001.

\end{document}